\documentclass[10pt,twocolumn,letterpaper]{article}

\usepackage{3dv}
\usepackage{times}
\usepackage{epsfig}
\usepackage{graphicx}
\usepackage{amsmath}
\usepackage{amssymb}

\usepackage{xcolor}
\usepackage[pagebackref=true,breaklinks=true,letterpaper=true,colorlinks,bookmarks=false]{hyperref}
\usepackage{multirow}
\usepackage{booktabs}
\usepackage{microtype}
\usepackage{stfloats}

\usepackage[pagebackref=true,breaklinks=true,letterpaper=true,colorlinks,bookmarks=false]{hyperref}

\threedvfinalcopy 

\def\threedvPaperID{30} 

\ifthreedvfinal\fi

\begin{document}

\newcommand{\misscite}{\textcolor{red}{[C]~}}
\newcommand{\datasetname}{Pixel-Face}
\newcommand{\methodname}{Pixel-3DM}
\newcommand{\missing}{\textcolor{red}{[missing]~}}
\newcommand{\rongyu}[1]{{\color{blue} (rongyu: {#1})}}
\newcommand{\ry}[1]{{\color{blue} (rongyu: {#1})}}
\newcommand{\zyx}[1]{{\color{red} (zyx: {#1})}}

\title{\datasetname: A Large-Scale, High-Resolution Benchmark \\ for 3D Face Reconstruction}

\author{
	Jiangjing Lyu$^{1}$ \hspace{9pt} Xiaobo Li$^{1}$ \hspace{9pt} Xiangyu Zhu$^{2}$ \hspace{9pt} Cheng Cheng$^{3}$\\
	\small{$^{1}$ Alibaba Inc. \hspace{9pt} $^{2}$Institute of Automation, Chinese Academic of Sciences
	\hspace{9pt} $^{3}$Chinese Academy of Sciences}\\
	{\tt\small \{ jiangjing.ljj, xiaobo.lixb\}@alibaba-inc.com \hspace{1pt} xiangyu.zhu@nlpr.ia.ac.cn  \hspace{1pt} ccheng2017@sinano.ac.cn}
}

\maketitle


\begin{abstract}
3D face reconstruction is a fundamental task that can facilitate numerous applications such as robust facial analysis and augmented reality. 
It is also a challenging task due to the lack of high-quality datasets that can fuel current deep learning-based methods.
However, existing datasets are limited in quantity, realisticity and diversity.
%
%
To circumvent these hurdles, we introduce \textbf{\datasetname}, a large-scale, high-resolution and diverse 3D face dataset with massive annotations. 
Specifically, \datasetname~contains 855 subjects 
aging from 18 to 80. Each subject has more than 20 samples with various expressions. Each sample is composed of high-resolution multi-view RGB images and 3D meshes with various expressions.
Moreover, we collect precise landmarks annotation and 3D registration result for each data. 
To demonstrate the advantages of \datasetname, we re-parameterize the 3D Morphable Model (3DMM) into \textbf{\methodname}~using the collected data.
We show that the obtained \methodname~is better in modeling a wide range of face shapes and expressions.
%
We also carefully benchmark existing 3D face reconstruction methods on our dataset. 
Moreover, \datasetname~serves as an effective training source. We observe that the performance of current face reconstruction models significantly improves both on existing benchmarks and \datasetname~after being fine-tuned using our newly collected data.
Extensive experiments demonstrate the effectiveness of \methodname~and the usefulness of \datasetname.
%

\end{abstract}


\section{Introduction}


\begin{figure*}[t]
	\begin{center}
		\includegraphics[width=1\linewidth]{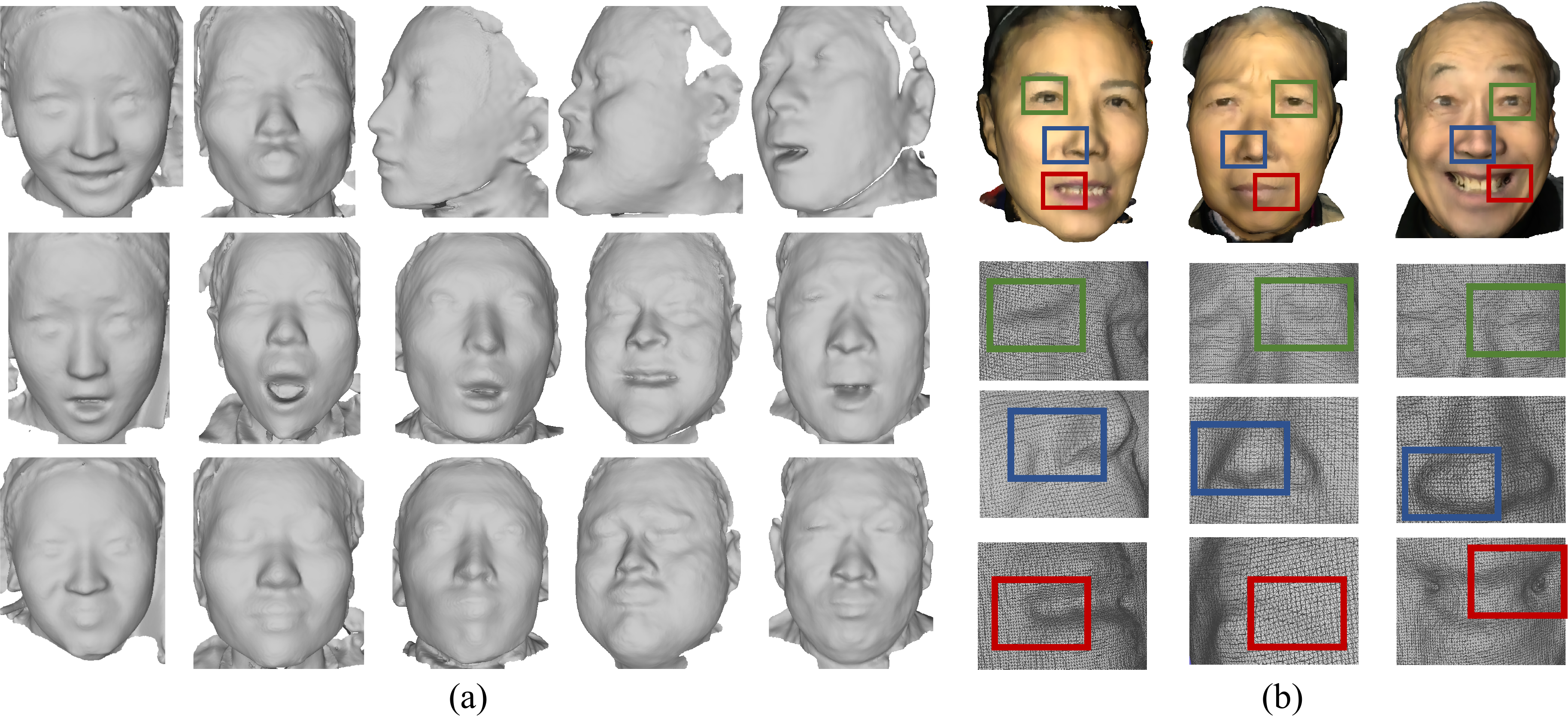}
	\end{center}
	\vspace{-5pt}
	\caption{\textbf{3D samples of {\datasetname}.} In (a), we show several typical 3D samples with different gender, age and expressions. In (b), we visualize the facial details preserved by high-resolution 3D meshes.}
	\label{fig:data_illustration}
\end{figure*}

Monocular 3D face reconstruction is one of the most fundamental tasks in computer vision~\cite{feng2018prn,Gecer_2019_CVPR,zhu2017face}. 
However, the research of 3D face analysis is obstructed by several inherent challenges.
First, obtaining ground-truth 3D annotations for in-the-wild images is both expensive and laborious. 
Firstly, sophisticated devices such as Kinect are used to capture raw 3D point clouds. Furthermore, obtaining multi-modality 3D data requires complicated processing that includes multi-views scanning, depth generating, landmarks annotation, point clouds fusion and 3D surface meshes generation. 
%
Second, current 3D face analysis methods majorly rely on a valid 3D Morphable Model to perform precise 3D face reconstruction. 3DMM, however, is sensitive to the quantity and quality of training data and can be easily affected by many factors such as age, gender and expression~\cite{dai20173d}.
Third, as the result of training on synthetic 3D face datasets such as 300W-LP~\cite{blanz1999morphable},
most state-of-the-arts 3D face reconstruction methods have limited capacity in representing real face shapes and expressions. 

In the past decade, although several authentic 3D face datasets have been released, they all have some non-negligible shortcomings.
Early datasets such as Bosphorus~\cite{savran2008bosphorus} only provide low-precision 3D meshes. 
BFM~\cite{paysan20093d} uses synthetic images which leads to the poor generalization ability of models trained on it.
Follow-up datasets overcome these shortcomings but they only provide limited annotations.
Texas-3D~\cite{gupta2010texas} only offers depth information. FWH~\cite{cao2013facewarehouse} provides results of 3DMM fitting instead of the original 3D meshes. The 3D annotations collected by MICC~\cite{Bagdanov2011} are not paired with the 2D images.
Recent works overcome the aforementioned drawbacks but they are limited in data diversity. 
3DFAW~\cite{krishnan20192nd} only collects data from 26 identities with fixed neutral expressions.
%
BP4D~\cite{zhang2014bp4d} only provides single-view 3D meshes.
The age distribution of FaceScape~\cite{yang2020facescape} lies major in 18 to 25.

In view of those shortcomings and aim to further push forward the research of 3D face reconstruction, we introduce \datasetname, a large-scale 3D face dataset with diverse samples and comprehensive annotations, as shown in Fig.~\ref{fig:data_illustration}.
Compared with existing datasets, our new benchmark has several appealing properties:
\textbf{1)}~\textit{Quantity.} -
\datasetname~contains a training set with 655 identities and an evaluation set with 200 identities. 
Each subject has over 20 images with paired 3D annotations under different expressions and multiple views.
\textbf{2)}~\textit{Quality.} - We use a  high-precision trinocular structured light system (Ainstec) and surface mesh generation method~\cite{burns2009centroidal} to obtain high-quality 3D meshes with resolution of $0.1292 \pm 0.0128 mm$~\cite{zuo2018phase}. 
%
Besides, we perform multi-view fusion and registration to get aligned meshes based on pre-defined templates, such as 3DMM~\cite{blanz1999morphable}.
\textbf{3)}~\textit{Diversity.}
The age of the subjects ranges from 18 to 80. Each subject has more than 10 expressions with 3D meshes and synchronized 2D images under multiple views.
\textbf{4)}~\textit{Availability.} - \datasetname~will be made publicly available to the research community.

To demonstrate the usefulness of {\datasetname}, we construct a new 3DMM, named {\methodname} and conduct extensive experiments to compare it with previous 3DMM models. Facilitated by the high-quality and diverse annotations provided by \datasetname, \methodname~surpasses all previous methods in representing more precise face shapes.
Comprehensive evaluation set provided by~\datasetname~enables us to rigorously benchmark the performance of existing 3D face reconstruction methods. 
The pitfalls of the methods trained by existing datasets are revealed.
They have limited capability of reconstructing real 3D faces with various shapes and challenging attributes such as exaggerated expressions or uncommon age.
We believe it is due to the domain gap between the authentic data and previous synthetic datasets these models were trained on.
Given this fact, we further finetune representative methods with the training set of \datasetname. After finetuning, the performance on both existing benchmarks and ours can be improved by 7\%-30\%, which demonstrates the effectiveness of \datasetname~as a pre-training source.

In summary, the contributions of this work are three-fold:
\textbf{1)} We build a large-scale 3D face dataset with carefully-collected training and evaluation sets.
The dataset is composed of multi-view, high-resolution and diverse 2D face images with paired high-quality 3D annotations.
\textbf{2)} We construct \methodname, a more expressive new 3DMM trained with massive diversely-distributed 3D face data. 
Comparison with previous 3DMMs illustrates the strengths of \methodname~in modeling face shapes and expressions. 
\textbf{3)} Third, we perform a comprehensive evaluation of existing methods on our benchmark and reveal several valuable observations.
We finetune representative methods with the training set of \datasetname. 
Experimental results demonstrate that the performance of current state-of-the-art methods can be significantly improved after finetuning.


\section{Related Work}

\subsection{3D Face Datasets}
3D face reconstruction can faciliate many tasks such as face animation~\cite{cao20133d,cao2016real} robust face recognition~\cite{cao2018pose,wright2008robust} and human motion capture~\cite{joo2018total,kanazawa2018end,rong2019delving,rong2020frankmocap,xiang2019monocular}.
Despite its important, 3D ground truth is unavailable for most in-the-wild 2D images.
The lack of paired 2D and 3D data hinders the training and evaluation of 3D face reconstruction methods. 
To alleviate this problem, 3DDFA~\cite{zhu2017face} builds a training dataset composed of 2D images and pseudo-3D meshes which are obtained from 3DMM fitting~\cite{romdhani2005estimating} and manually adjusting.
Another evaluation dataset named AFLW2000-3D~\cite{Huang2012a} using the same method.
The ambiguous typology of synthetic 3D data limits its capability of representing intricate faces.
Afterwards, some authentic 3D evaluation datasets are released, such as the Florence 2D/3D hybrid face dataset (MICC)~\cite{Bagdanov2011}, ``Not quite in-the-Wild'' dataset (NoW)~\cite{sanyal2019learning}, BP4D dataset~\cite{zhang2014bp4d} and FaceScape dataset~\cite{yang2020facescape}.
However, these datasets often confront the problems of lacking diversity of identities and attributes.
As a result, it is still a challenge to develop 3D face reconstruction methods to generate realistic 3D face meshes.
A large-scale, high-resolution and multi-modality 3D face dataset with affluent annotation is required to deal with the above problems. 

\subsection{3D Face Models}
A 3DMM is composed of facial shape and expression models. 
Given a face image, the corresponding 3D mesh can be reconstructed by fitting the coefficients of a 3DMM model.
%
%
Pascal et.al~\cite{paysan20093d} constructs the Basel Face Model~(BFM) from 200 registered face meshes with neutral expressions. 
Thomaset.al~\cite{gerig2018morphable} updates the BFM model by adopting 100 additional individuals from Binghamton University 3D Facial Expression Database (BU-3DFE)~\cite{yin20063d}. 
FaceWareHouse~\cite{cao2013facewarehouse} is an elaborate expression model constructed by 150 subjects with over 20 expressions. 
FLAME~\cite{li2017learning} is constructed from CAESAR dataset~\cite{robinette2002civilian} with low-resolution 3D face meshes. 
In general, most of the current 3DMMs are constructed from a small 3D dataset with less than 200 subjects. They tend to suffer from low precision and monotonous expressions. %
As a result, the generalization capability of these 3DMMs in real applications cannot be guaranteed. 
Taking advantage of \datasetname, we construct a new 3DMM, \methodname, which is more accurate and reliable in face representation.


\section{The \datasetname~Dataset}


\begin{figure*}[t]
	\begin{center}
		\includegraphics[width=1\linewidth]{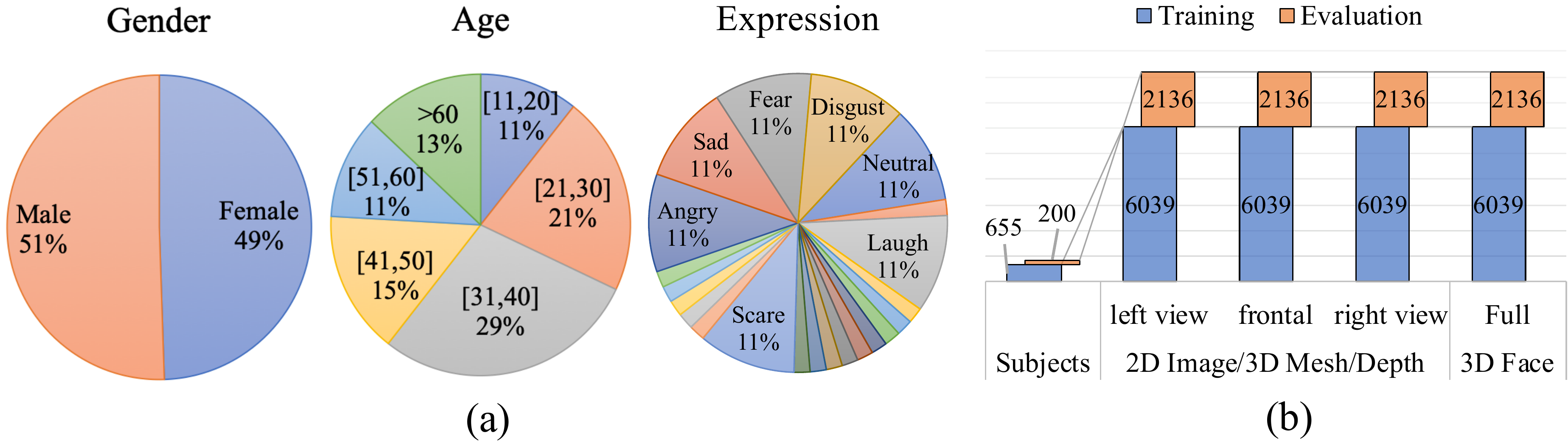}
	\end{center}
	\vspace{-5pt}
	\caption{\textbf{Overview of {\datasetname}.} In (a), we show gender, age and expression distribution of the {\datasetname}. {\datasetname} covers balanced gender, wide-range ages and massive expressions. In (b),  we show other statistical information including training / evaluation split and amount of data.}
	\label{fig:data_distribution}
\end{figure*}


\begin{table*}[t] \centering
\caption{Comparing \datasetname~with other authentic 3D Face datasets. The \datasetname~has advantages in most aspects. Lms., Exp. and Vert. are abbreviations for the annotation number of facial landmarks, categories of expressions and number of vertices, respectively.
}
\setlength\tabcolsep{5pt}
\begin{tabular}{lllllllll}
\toprule
Dataset                     & Sub. Num & Image Num & 3D Mesh Num   & Lms. Num. & Exp. Num.         & View & Camera    & Vert. Num \\
\hline
Bosphorus~\cite{savran2008bosphorus}                   & 105 & 4666 & 4666     & 24   & \textbf{35}           & Single         & Mega      & 35k     \\
BFM~\cite{paysan20093d}                      & 200 & synthetic & 200 & 68   & Neutral         & Single         & ABW-3D    & 50k     \\
FWH~\cite{cao2013facewarehouse}& 150 & 3000 & 3DMM & 74 & 20 & Single & Kinect v1& 20k\\
MICC~\cite{Bagdanov2011}     & 53 & 53 & 203         & 51   & \textless{}5 & Single         & 3dMD      & 40k     \\
3DFAW~\cite{krishnan20192nd}   & 26 & 26 & 26          & 51   & Neutral         & Single         & DI3D      & 20k     \\
BP4D~\cite{zhang2014bp4d}     & 41 & 328 & 328     & 84   & 8            & Single         & 3DMD      & 70k     \\
FaceScape~\cite{yang2020facescape} & 359 & 400,000 & 7120 &106& 20& Multi & DSLR & 2m\\
\midrule
\datasetname & \textbf{855} & \textbf{24,525} & \textbf{24,525}  & \textbf{106}  & 22           & \textbf{Multi}        & Ainstec        & 100k\\
\bottomrule
\end{tabular}
\label{tab:pub_dataset}
\vspace{0.1 cm}
\end{table*}


\begin{figure*}[t]
	\begin{center}
		\includegraphics[width=1\linewidth]{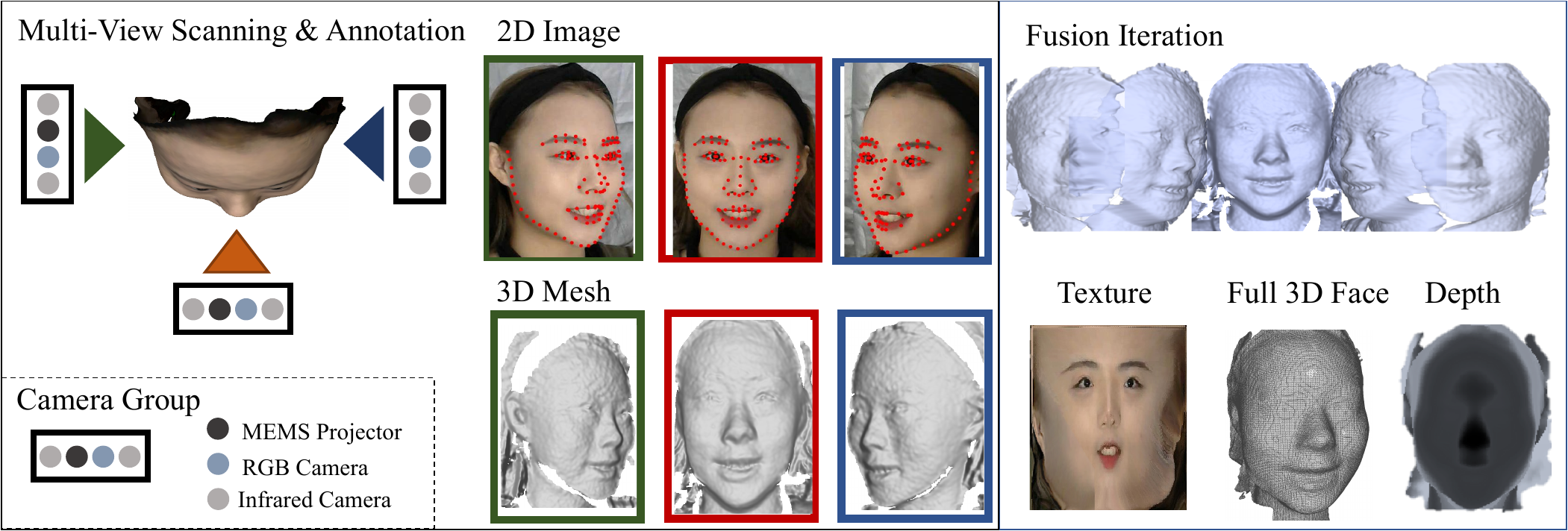}
	\end{center}
	\caption{\textbf{Data collecting pipeline.} The pipeline is composed of multi-view scanning, facial landmarks annotation, texture mapping, multi-view fusion, surface mesh generation and depth generation. High-resolution, multi-modality data and comprehensive annotations are obtained by our elaborately designed pipeline.
	}
	\label{fig:pipeline}
\end{figure*}

We contribute {\datasetname}, a large-scale 3D face dataset with affluent annotations.
{\datasetname} has several appealing properties. First, it is the largest, high-fidelity 3D face dataset. 
\datasetname~contains over 24,000 multi-modality samples collected from 855 subjects under different views. Each data sample contains both RGB images and 3D meshes with corresponding face landmarks.
The full 3D faces obtained from full-view fusion are also provided.
The statistical information of {\datasetname} is shown in Fig.~\ref{fig:data_distribution} (b).
The high-resolution 3D meshes have advantages on preserving details of authentic faces, as shown in Fig.~\ref{fig:data_illustration} (b).
Second, \datasetname~offers manually annotated facial landmarks for each face mesh. 
These landmarks can aid the tasks including multi-view fusion, 3D mesh registration and 3D face reconstruction.
Third, these subjects cover balanced gender, wide-range age and various expression distribution.
Distribution of different attributes is shown in Fig.~\ref{fig:data_distribution} (a).
More details of expressions are provided in the supplementary. 
Comparisons between different datasets shown in Tab.~\ref{tab:pub_dataset} reveals that \datasetname~surpasses the existing datasets in terms of scale, quality of annotations and diversity of views.

\subsection{Data Acquisition}
\label{sec:data_acquisition}
Fig.~\ref{fig:pipeline} demonstrates the pipeline of collecting 3D face data. 
From 855 diverse subjects, we collect over 24,000 raw 3D point clouds with high-resolution of $0.1292\pm0.0128$ $mm$ 
using a self-customized trinocular structured light system~\cite{xue20193d}.

\noindent
\textbf{Multi-View Scanning.}
To avoid collected data being corrupted by self-occlusion, we set three camera-groups surrounding subjects' head to cover 270 degrees.
Following similar settings in~\cite{booth20163d}, each camera group contains one RGB camera, one MEMS projector, and two infrared cameras.  
We accomplish the whole scanning process using N-step phase shifting~\cite{zhang2018high}, which is a state-of-the-art 3D scanning method to capture 3D point clouds with pixel-wise resolution.
This method alleviates the influence of varied surface reflectivity effectively.
After scanning, we can simultaneously acquire 2D images and corresponding 3D point clouds. The average processing time for each sample is less than 300 ms. 

\noindent
\textbf{3D Face Landmarks Annotation.}
To get 3D facial landmarks, directly annotating 3D landmarks on the raw point clouds is time-costing. Therefore, we apply a retrieval-based method.
Firstly, we manually annotate 106 2D facial landmarks for each 2D image. The facial landmarks are defined the same as in~\cite{liu2019grand}.
For each vertex in the point clouds, we apply texture mapping method~\cite{zhou2014color} to calculate the corresponding coordinates on the 2D images.
To find the corresponding 3D landmark for each annotated 2D facial landmark, we calculate the distance between 2D landmarks and the projected 2D coordinates of each 3D vertex in point clouds. The nearest 3D vertex is retrieved as the corresponding 3D landmark.

\noindent
\textbf{Multi-View Fusion.}
To get full-view point clouds, we employ an improved coarse-to-fine Iterative Closest Point (ICP) \cite{besl1992method} to fuse the captured 3D point clouds in three views (left, middle, and right). 
To get the coarse results, we use the corresponding 3D landmarks to calculate the transformation relationship between point clouds in different views.
We set the middle mesh as the pivot and align the left and right point clouds to the pivot by calculating the rigid transformation matrix coarsely.
The coarse fusion results have limitations in the smoothness of a surface, seamless integration between edges and precision of details.
Therefore, we further refine the fusion results by iteratively calculating the transformation matrix for each vertex~\cite{besl1992method}.
Finally, we merge the overlapped vertices and omit the isolated vertices.
After fusion, we obtain full-view 3D point clouds and complete 3D landmarks.
The corresponding 3D meshes of point clouds are obtained by Centroidal Voronoi Tessellation (CVT) based method~\cite{burns2009centroidal}. 
The resulting meshes serve as the full 3D faces.

\noindent
\textbf{3D Point Cloud to Depth.}
To enable more tasks such as monocular depth prediction, we provide depth images for each sample.
We use structured light and triangulation~\cite{ladicky2017point} to calculate depth value for each vertex in 3D point clouds.
Each vertex has the corresponding coordinate in the 2D image so that the depth information can be projected to 2D depth image.
In this way, a depth image for each 3D point cloud is obtained.

\subsection{Rich Semantic Annotations}
Besides the 3D landmarks mentioned before, the \datasetname~offers semantic annotations~\cite{liu2015deep} including gender, age and expression.
To obtain these annotations, we first collect information such as gender and age from each subject. 
Then we demand each subject to perform 22 pre-defined expressions adopted from FaceWareHouse~\cite{savran2008bosphorus}.
The distributions of gender, age and expression are shown in Fig.~\ref{fig:data_distribution}. 
The \datasetname~dataset has both balanced gender and wide-range age distribution.
Besides, each subject contains rich expressions.

%
\section{Construct \methodname}
To demonstrate the usefulness of \datasetname~and facilitate future research, we use the obtained face meshes to construct a new 3DMM, named \methodname.
It contains both facial shape models and meticulous expression models. 
To obtain the new 3DMM, we first register the initial 3D meshes to a 3D template by an improved two-stage algorithm.
Then we use the registration results to calculate PCA bases of 3DMM. 
Fig.~\ref{fig:3dmm_illustration} (a) demonstrates the process of registration and show an example of registration result. Fig.~\ref{fig:3dmm_illustration} (b) visualizes that the {\methodname} can be driven by face shape and expressions bases flexibly.

\subsection{3D Face Registration}
3D face registration aims to align the arbitrary 3D meshes with some pre-defined mesh template, 
so that the registration results have consistent topology.
In this subsection, we first give a brief introduction to 3DMM and then discuss the methods used for registration.

\noindent \textbf{3DMM.}
3D Morphable Model (3DMM)~\cite{booth20163d} represents any 3D face mesh $M$ as Eq.~\ref{eq:3DMM}.

\begin{equation} 
\label{eq:3DMM}
	M = \bar{M} + \sum{\alpha_i U_i} + \sum{\beta_j E_j},
\end{equation}

\noindent
$\bar{M}$ is the mean shape. $U$ and $E$ refer to the orthonormal bases matrix whose columns are the shape and expression eigenvectors computed from PCA.
The $\alpha$ is the shape coefficients and the $\beta$ is expression coefficients. 
The combination of these terms determines a specific instance under the given 3DMM.


\begin{figure}[t]
	\begin{center}
		\includegraphics[width=1\linewidth]{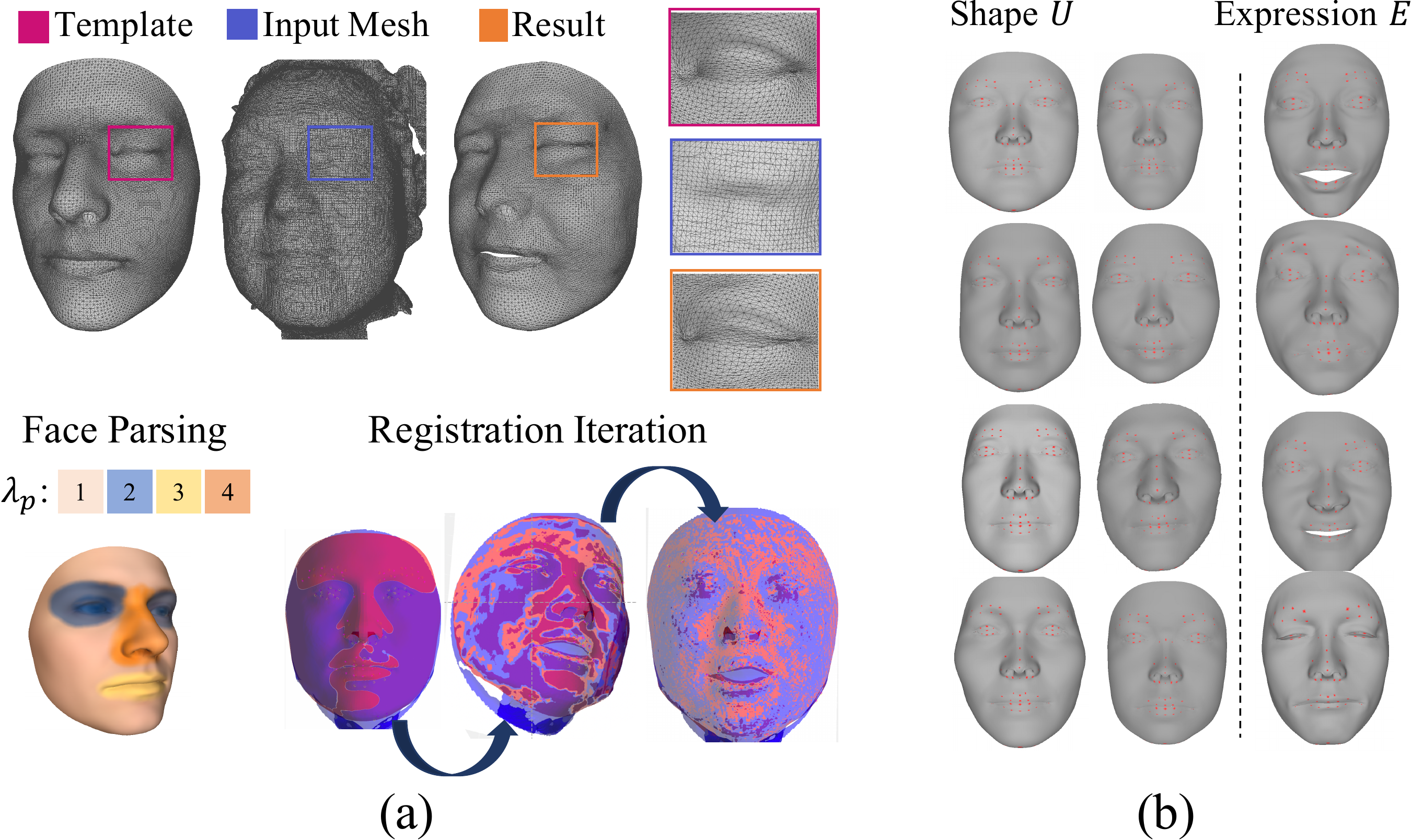}
	\end{center}
	\vspace{-16pt}
	\caption{\textbf{Overview of {\methodname}.} At the top of (a), we show an example of an input mesh and its registration result which preserves the facial shape and expression information while sharing the consistent topology with the template. Details are shown in the colored box. The left bottom of (a) illuminates the face parsing and spatial-varying $\lambda_{p}$ defined in Eq.~\ref{eq:NICP}. The registration iteration is shown at the right bottom of (a). In (b), we show several typical shape and expression bases of \methodname.
	} 
	\label{fig:3dmm_illustration}
\end{figure}

\noindent \textbf{Registration.}
To register the obtained 3D face meshes to 3DMM, we adopt a two-stage algorithm that combines the strengths of two popular registration methods, ICP~\cite{besl1992method,chen1992object} and NICP~\cite{amberg2007optimal}.
The goal of 3D face registration is to fit template mesh $\mathcal{S}=\left(\mathcal{V}_{S}, \mathcal{F}_{S}\right)$ to template-free mesh $\mathcal{T}=\left(\mathcal{V}_{T}, \mathcal{F}_{T}\right)$.
$\mathcal{V}$ and $\mathcal{F}$ refers to 3D vertices and faces, respectively.
After registration $\boldsymbol{R}$, the $\mathcal{T}$ can be represented in the form of $\mathcal{S}$ as shown in Eq.~\ref{eq:register}:

\begin{equation}
 \label{eq:register}
 \mathcal{T} \approx \boldsymbol{R}(\mathcal{S})=\left(\boldsymbol{R}\left(\mathcal{V}_{S}\right), \mathcal{F}_{S}\right),
\end{equation}
where $\boldsymbol{R}\left(\mathcal{V}_{S}\right)$ refers to relocated vertices and $\mathcal{F}_{S}$ refers to the mesh faces defined by $\boldsymbol{S}$.
A advanced registration methods should guarantee the $\boldsymbol{R}(\mathcal{S})$ be close enough to $\mathcal{T}$. Specifically, $\boldsymbol{R}(\mathcal{S})$ is required to represent the facial expression, head pose and identity of $\mathcal{T}$ precisely and robust to challenging cases such as exaggerated expressions or data missing.

In the first stage of registration, we apply the same method used in section~\ref{sec:data_acquisition} to fuse point clouds.
To get the ICP result $\boldsymbol{R}_{icp}(\mathcal{S})$, we first estimate the transformation matrix between the landmarks of $\mathcal{T}$ and $\mathcal{S}$, and then use the obtained transformation matrices to transform $\mathcal{T}$ to $\mathcal{S}$.
Since the ICP-based method only generates coarse meshes and cannot handle subtle face details,
we further deploy a spatial-varying NICP as the second stage to refine the detail meshes.

The conventional NICP-based method~\cite{amberg2007optimal} does not contain valid stiffness contraint.
As a result, the transformation matrix calculated by NICP is less constrained and prone to dislocations of points.
For example, points of the nose may move to the cheek and different points may occupy the same position. 
To resolve this problem, we adopt a spatial varying deformation method.
%
We manually segment the face to several parts $P$, according to both semantic information and spatial location. Each part has the corresponding surface $\mathcal{T}_{p}$.
Then we calculate transformation matrix of each face vertex.
The cost function is defined as Eq.~\ref{eq:NICP}.

\begin{equation}
\label{eq:NICP}
	\sum_{p \in P}\left(\sum_{i \in p} \left(w_{p}^{i} \operatorname{dist}\left(\mathcal{T}_{p}, X_{p}^{i} v_{p}^{i}\right) +\lambda_{p}\sum_{\{i, j\} \in \mathcal{E}}\left\|X_{p}^{i}-X_{p}^{j}\right\| \right)\right) 
\end{equation}

\noindent
$v_{p}^{i}$ refers to vertex in $\boldsymbol{R}_{icp}(\mathcal{S})$ and $X_{p}^{i}$ is the corresponding transformation matrix.
The first term affects registration accuracy, 
$w_p$ refers to the importance weight of each vertex(we set it to $1$ in practice).
We calculate the euclidean distance of one vertex in $\boldsymbol{R}_{icp}(\mathcal{S})$ to the closest counterpart in $\mathcal{T}$.
%
This distance is marked as $\operatorname{dist}(\mathcal{T}, v)$.
The second term is the stiffness regularization. 

$\mathcal{E}$ refers to a small region. In practice, we set it to be a unit sphere.
$\lambda_p$ is the trade-off weight to balance the flexibility and stiffness of deformation.
Higher $\lambda_p$ corresponds to a stiffer restriction. 
Since different parts of faces have specific surface curvature, the $\lambda_p$ is set to specific values for each part.
For example, the surface of cheek is smoother than the nose, so the transformation of points in the nose tends to be intenser and leads to more dislocations. 
Part division of faces and the corresponding value of $\lambda_p$ is shown in Fig.~\ref{fig:3dmm_illustration} (a). After minimizing the Eq.~\ref{eq:NICP} using least square algorithm, we obtain optimized $X_p^i$ for each face vertex $v_p^i$. In the end, each vertex is transformed accordingly.

\subsection{\methodname}
We follow the general process of constructing 3D morphable model~\cite{blanz1999morphable} to build {\methodname}.
%
%
%
%
%
%
%
%
We concatenate over 600 registration results with the neutral expressions as facial shape matrix.
$\bar{M}$ is set to be the mean of those facial shape matrix. 
The shape model $U$ is composed of 199 PCA components covering more than 99\% of the variance observed in the facial shape matrix.
To obtain expression model $E$, we use over 6000 registration results with various expressions. 
For each sample, we compute its residual to the corresponding registration result with neutral expression and then concatenate these residuals to form the expression residual matrix.  
The expression model $E$ is composed of 99 components explaining more than 99\% of the variance observed in the expression residual matrix.


\section{Experiments}
\subsection{Benchmarks}
We build benchmarks out of \datasetname~for evaluating 3D face reconstruction methods. 
The task of 3D face reconstruction is to predict the 3D mesh taken the 2D image as input.
There are a bunch of previous works~\cite{savran2008bosphorus,tran2018extreme,zhu2017face} that focus on 3D face reconstruction.
In this paper, we choose the three most representative methods and evaluate their performance on the newly-obtained {\datasetname}.
Detailed evaluating results are reported in the section~\ref{sec:result}.

\noindent
\textbf{3DMM Fitting.}
With the optimization object based on facial landmarks~\cite{blanz1999morphable} and the 3DMM assumption in Eq.~\ref{eq:3DMM}, 3DMM Fitting method~\cite{aldrian2012inverse} formulates 3D face modeling as an optimization problem to fits 3DMM coefficients.
Since these optimization-based methods do not require training, we can directly apply them to our \methodname.

\noindent
\textbf{Coefficient Regression Model.}
Different from 3DMM fitting, these methods~\cite{Gecer_2019_CVPR,RingNet:CVPR:2019,zhu2017face} use deep convolutional neural networks (DCNN) to directly regress model coefficients. 
These models should be re-trained if the 3D bases change. 
In our experiment, we evaluate two methods on {\datasetname}, namely, 3DDFA~\cite{zhu2017face} and RingNet~\cite{RingNet:CVPR:2019}.

\noindent
\textbf{Dense Map Model.} 
These methods~\cite{feng2018prn,tran2018extreme} directly predict 3D dense reconstructions, such as UV position map~\cite{feng2018prn} from input 2D images. 
The DCNN often serves as the backbone.
We evaluate PRNet~\cite{feng2018prn} in our experiment.

\subsection{Experimental Settings}\label{sec:setting}
\noindent
\textbf{Data.}
We mainly use the newly-built {\datasetname} to conduct the experiments.
{\datasetname}~is composed of 855 subjects with more than 24,000 multi-view samples.
Each sample is composed of a high-resolution RGB image, a high-quality 3D mesh and 3D landmark annotations.
We use 75\% of \datasetname~for training and the rest for evaluation (validation+test) as shown in Fig.~\ref{fig:data_distribution} (b).
We pre-train {\methodname} and fine-tune other methods by the training set. 
Besides the evaluation set of \datasetname, a subset of BP4D dataset~\cite{zhang2014bp4d} are also used in section~\ref{sec:result}.
The BP4D dataset is a 3D expression dataset containing 41 identities each of which offers about 8 tasks of expression.
There are paired 2D/3D scanning sequence for each expression task.
To remove redundant information from adjacent frames, we randomly sample one pair of 2D/3D data from each sequence.
After sampling, a subset containing 328 2D/3D pairs of data with different expressions from 41 identities is obtained. 


\begin{figure*}[t]
	\centering
	\includegraphics[width=1\linewidth]{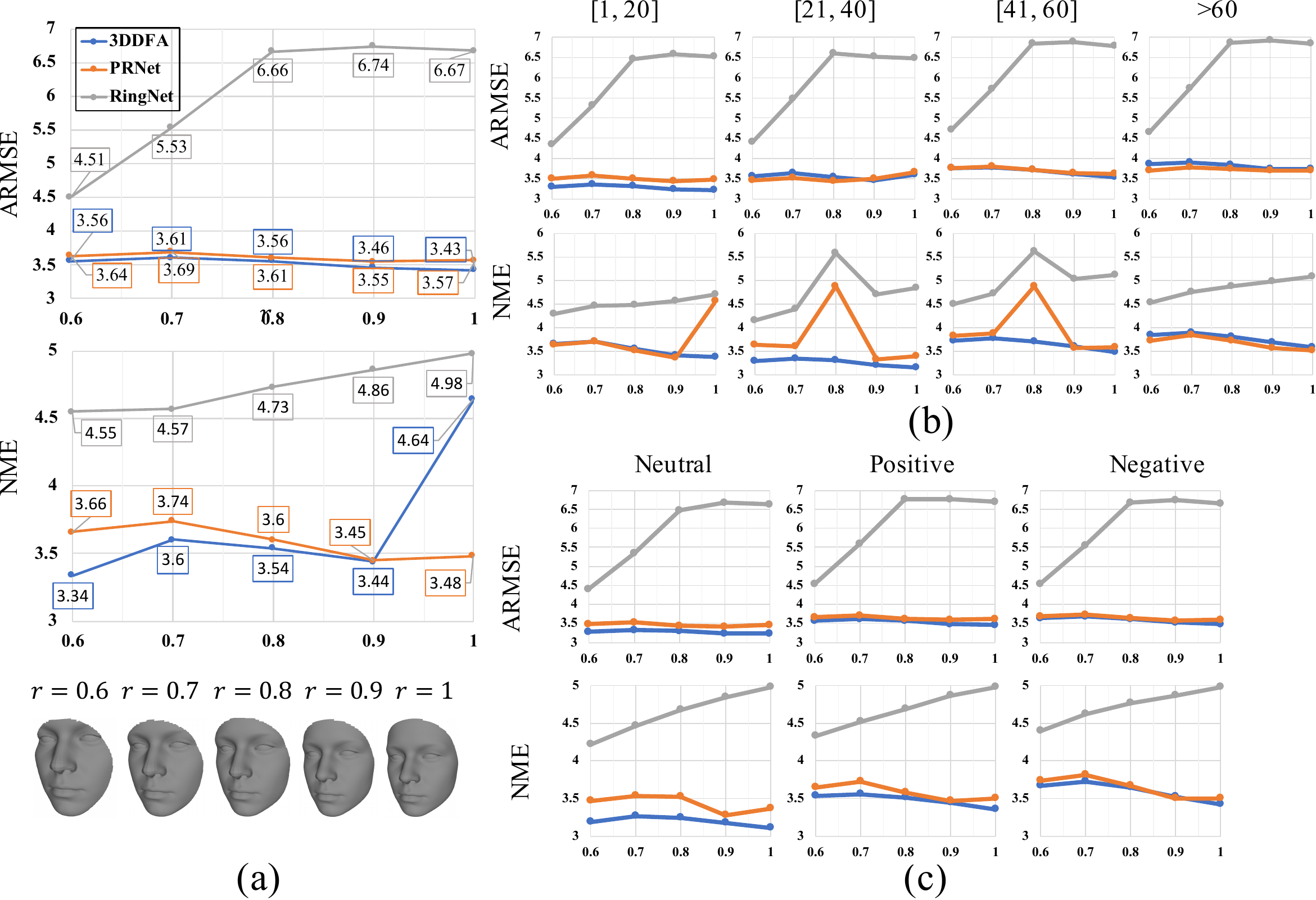}
	\vspace{-9pt}
	\caption{
		In (a), we plot the curve of overall ARMSE and NME scores of three 3D methods, namely 3DDFA~\cite{zhu2017face} (\textcolor{blue}{blue}), PRNet~\cite{feng2018prn} (\textcolor{orange}{orange}) and RingNet~\cite{RingNet:CVPR:2019} (\textcolor{gray}{grey}). The visualization of faces under different crop radius $r$ is shown in the bottom of (a). In (b), we plot the curves of ARMSE and NME under different age regions. In (c), we plot the curves of ARMSE and NME under different expression regions. The results show that the neutral expression and common ages are handled more easily than other sub-groups, revealing the limitations of the previous 3D face datasets.}
	\vskip -0.1cm
	\label{fig:err_age_exp}
\end{figure*}

\noindent
\textbf{Evaluation Metrics.}
We use NME and ARMSE as the evaluation metrics in our experiment.
The Normalized Mean Error (NME) is defined as the average of landmark errors normalized by the bounding box sizes~\cite{jourabloo2015pose}.
The Average Root Mean Square Error (ARMSE)~\cite{jeni2019second} is employed to evaluate the similarity between reconstructed 3D meshes and ground truth meshes.
Following the setting of $2nd$ 3DFAW challenge~\cite{jeni2019second}, we first normalize the interocular distance of ground truth to 1.
Then we align the reconstructed 3D meshes to the ground truth by facial landmarks. 
The origin is set to be the nose tip.
Given a crop radius which is marked as $r$, we discard vertices whose distance between nose tip is higher than $r$.
The ARMSE computes the closet point-to-mesh distance between the ground-truth and reconstructed 3D meshes and vice versa.
In our experiment, $r$ ranges from $0.6$ to $1.0$.


\begin{table}[t] \centering  
	\caption{We quantitatively compare {\methodname} with BFM17~\cite{gerig2018morphable} and FWH~\cite{cao2013facewarehouse} on evaluation set of {\datasetname}, under crop radius $r = 0.6$. The {\methodname} surpasses BFM17 and FHW in both ARMSE and NME.
	}

	\vskip 0.2 cm
	\setlength\tabcolsep{5pt}
	\begin{tabular}{ccc}
		\toprule
		Evaluation Metrics            $\rightarrow$  &   \multirow{2}{*}{NME} & \multirow{2}{*}{ARMSE}\\
		 3DMM Basis $\downarrow$    \\
		\hline
		 BFM17~\cite{gerig2018morphable} + FHW~\cite{cao2013facewarehouse}	&	$ 5.04$ &	$ 4.70$ \\
		 BFM17~\cite{gerig2018morphable} + \methodname ~Exp & $4.56$ & $4.77$ \\
		 \methodname~ Shape + FHW~\cite{cao2013facewarehouse} & $4.04$ & $3.97$ \\
		 
		 \midrule
		 \methodname & \textbf{3.84} & \textbf{3.80} \\
		\bottomrule
	\end{tabular}	
	\label{tab:3DMM}
\end{table}


\begin{table*}[t] \centering

\caption{We quantitatively evaluate the performances of PRNet~\cite{feng2018prn} with and without finetuning on our data. The results show that our data develop reconstructing capability of the PRNet both on the subset of BP4D~\cite{zhang2014bp4d} and the validation set of {\datasetname}.}

\setlength\tabcolsep{2pt}
\begin{tabular}{ccccccccccccccccccccc}
\toprule
Evaluation $\rightarrow$ & \multicolumn{10}{c}{NME} & \multicolumn{10}{c}{ARMSE} \\ \cline{2-21} 
Dataset $\downarrow$ & \multicolumn{5}{c}{w/o finetune} & \multicolumn{5}{c}{with finetune} & \multicolumn{5}{c}{w/o finetune} & \multicolumn{5}{c}{with finetune} \\ 
 $r$ &0.6 & 0.7 & 0.8 & 0.9 & 1& 0.6 & 0.7 & 0.8 & 0.9 & 1 & 0.6 & 0.7 & 0.8 & 0.9 & 1& 0.6 & 0.7 & 0.8 & 0.9 & 1\\
\hline
\datasetname &3.66 & 3.74 & 3.60 & 3.45 & 3.48 & \textbf{2.49} & \textbf{2.47} & \textbf{2.45} & \textbf{2.54} & \textbf{2.76} & 3.64 & 3.69 & 3.61 & 3.55 & 3.57 & \textbf{2.48} & \textbf{2.46} & \textbf{2.42} & \textbf{2.52} & \textbf{2.75}\\
BP4D~\cite{zhang2014bp4d} &2.44& 2.33& 2.27& 2.29& 2.51& \textbf{2.07}& \textbf{2.13}& \textbf{2.16}& \textbf{2.20}& \textbf{2.42} & 2.42 & 2.31 & 2.25 & 2.28 & 2.50 & \textbf{2.07} & \textbf{2.11} & \textbf{2.14} & \textbf{2.18} & \textbf{2.40}\\
\bottomrule

\end{tabular}
\label{tab:finetune}
\end{table*}
\vskip -0.2cm

\subsection{Results}\label{sec:result}
This section provides qualitative and quantitative evaluations of different methods on our benchmarks.
We firstly demonstrate the modeling capability of {\methodname} by comparing it with other representative 3DMMs on the evaluation set of {\methodname}, using the same optimization method.
Then we benchmark state-of-the-art reconstruction models on {\datasetname}.
Finally, to demonstrate the generalization ability of {\datasetname}, we finetune the PRNet~\cite{feng2018prn} with training set of {\datasetname} and evaluate both on the evaluation of {\datasetname} and a subset of BP4D~\cite{zhang2014bp4d}.

\noindent
\textbf{The Superiority of~\methodname.}
To compare the model capability between {\methodname} and previous 3DMMs, we apply the same 3DMM fitting method~\cite{faggian20083d} for different 3DMMs and verify their effectiveness on {\datasetname}. 
Specifically, we compare {\methodname} with face shape bases from BFM17~\cite{gerig2018morphable} and expression bases from FWH~\cite{cao2013facewarehouse}.
The results are listed in Tab.~\ref{tab:3DMM}. It is observed that \methodname~consistently outperforms BFM~\cite{paysan20093d} and FWH~\cite{cao2013facewarehouse} on both shape and expression modeling. 
The superior property of {\datasetname} validates the modeling capability of {\methodname}.


\begin{figure}[htb]
	\begin{center}
		\includegraphics[width=1\linewidth]{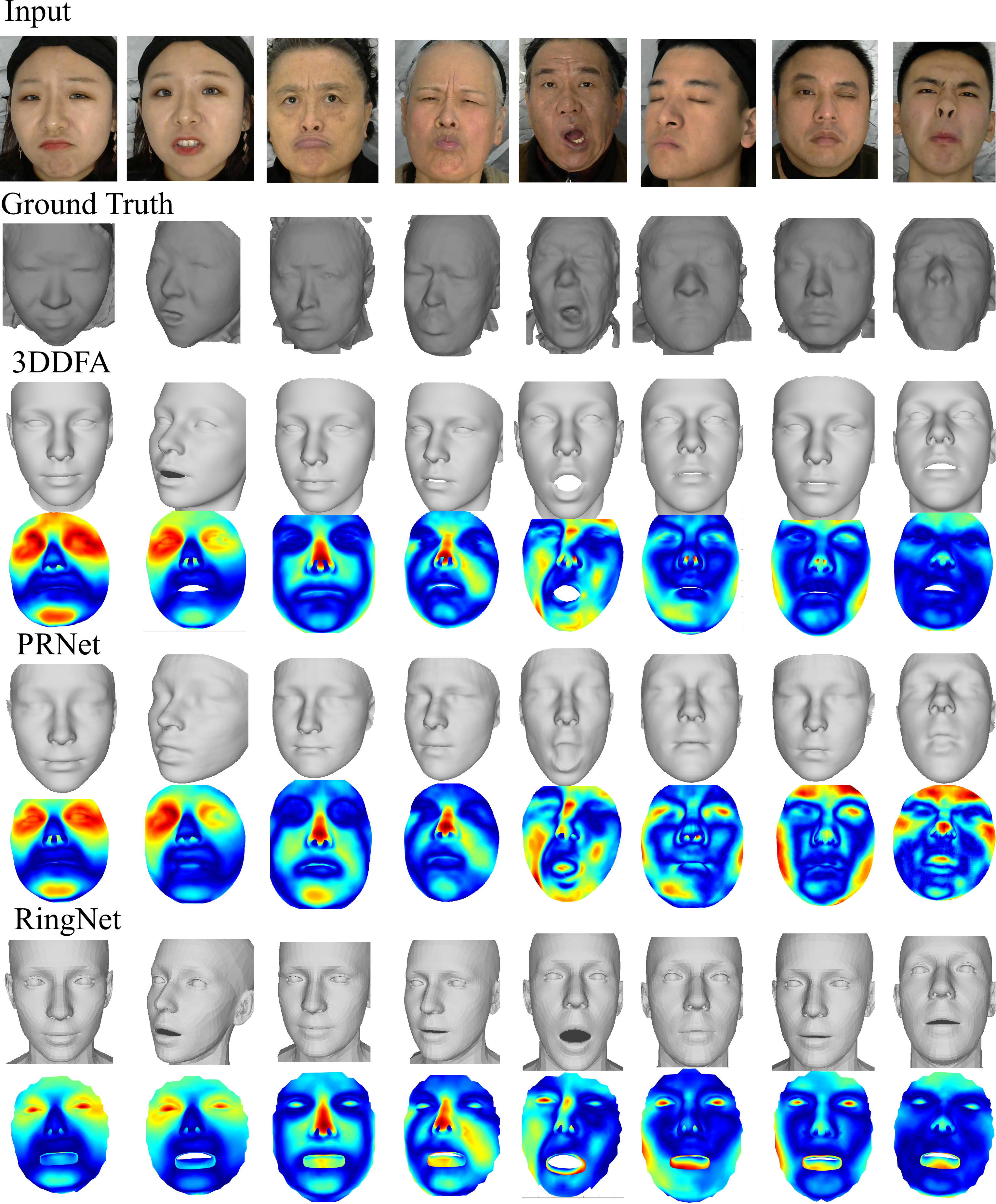}
	\end{center}
	\vspace{-10pt}
	\caption{We qualitatively compare 3DDFA~\cite{zhu2017face}, PRNet~\cite{feng2018prn} and RingNet~\cite{RingNet:CVPR:2019} on the evaluation set of {\methodname}, showing both the reconstructed 3D meshes and error maps. The results demonstrate that previous models trained with synthetic data tend to predict mean shape and neutral expressions and fall short in modeling authentic face shapes and various expressions.}
	\vspace{-10 pt}
	\label{fig:2d_recons}
\end{figure}


\begin{figure}[htb]
	\begin{center}		\includegraphics[width=1\linewidth]{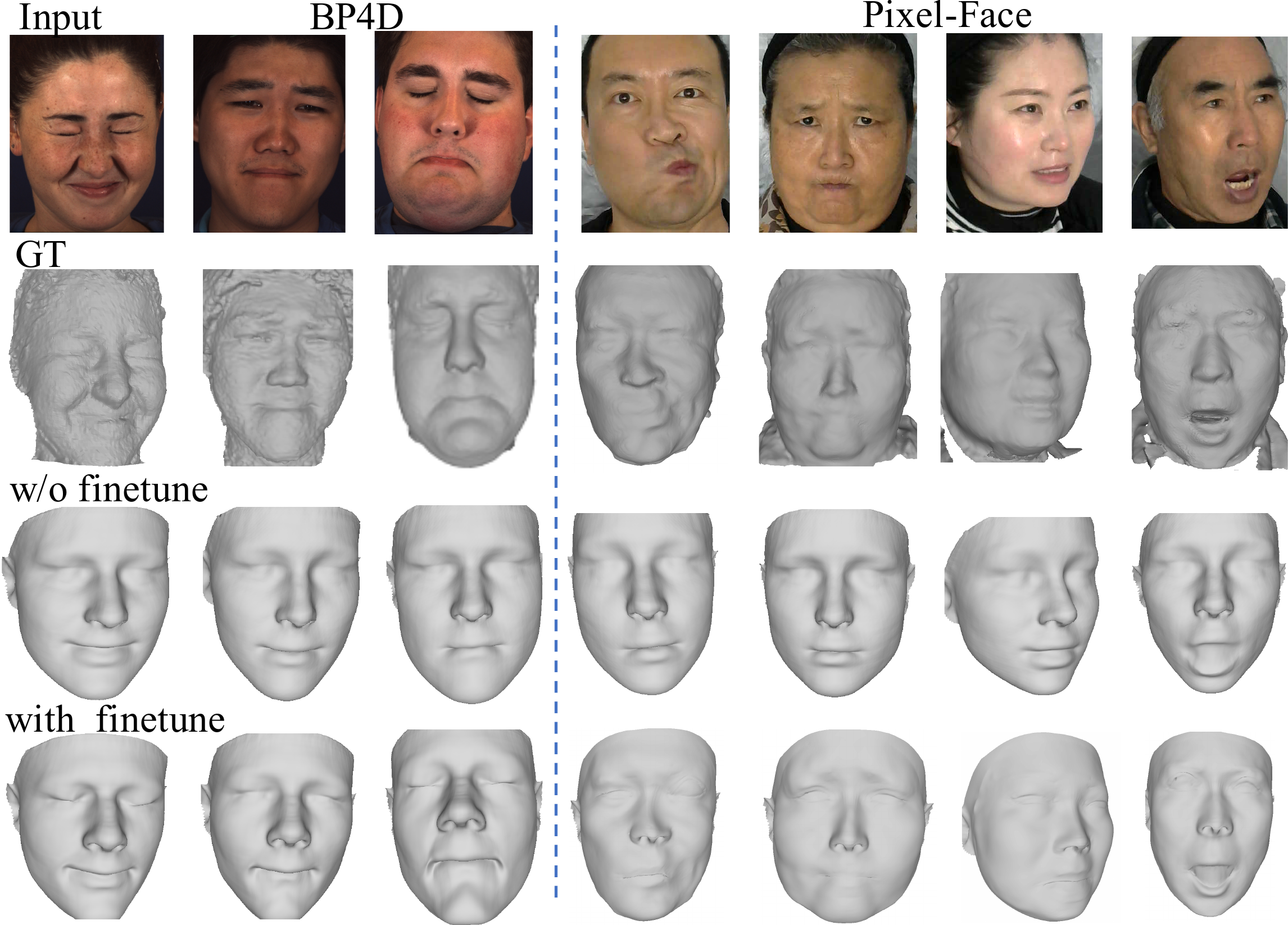}
	\end{center}
	\vspace{-5pt}
	\caption{
	We qualitatively compare performance of PRNet~\cite{feng2018prn} with and without finetuning on our dataset. The results show that our data can effectively promote the models' capacity in modeling various face shapes and expressions.}
	\vspace{-12pt}
	\label{fig:finetune}
\end{figure}

\noindent
\textbf{Benchmarking Results on~\datasetname.}
We evaluate several state-of-the-art methods on our \datasetname, including 3DDFA~\cite{zhu2017face}, PRNet~\cite{feng2018prn} and RingNet~\cite{RingNet2019}. 
We take NMS and ARMSE as metrics and evaluate different subsets of {~\datasetname} divided by expression and age.
Limited by the space, we roughly divided 22 expressions into three categories which are neutral, positive and negative in terms of emotion. We also split ages into 4 non-overlapping subsets.
Fig.~\ref{fig:err_age_exp} summarizes the performance of different methods on different expression and age subsets.
Fig.~\ref{fig:2d_recons} shows some qualitative results.
Several valuable observations are revealed from the experiment results: 
1) Although some synthetic 3D dataset~\cite{zhu2017face} has verisimilar 3D faces and diverse attributes, 
the models trained on them have limited capability of modeling real 3D faces.
2) The neutral expression and common ages are handled more easily than the uncommon ones.

\noindent
\textbf{\datasetname~as An Effective Training Source.}
We are curious about how well our data can promote other methods' performances on both \datasetname~and other previous datasets.
In addition to the evaluation set of \datasetname, we use the subset of the BP4D dataset~\cite{zhang2014bp4d} with 328 2D/3D pairs of data with different expressions. 
We finetune PRNet with the training set of \datasetname. The 3D ground-truth for each 2D image is generated by the preprocessing method provided by PRNet.
Tab.~\ref{tab:finetune} lists the experiment results of models with and without finetuning on {\datasetname}. 
Fig.~\ref{fig:finetune} shows the qualitative results.
The experiment results show that PRNet generates better aligned 3D meshes after finetuned on our dataset.
Using our data can further improve the performance of PRNet on both face shape and expression representing.
Specifically,  ARMSE drops by 30\% and NME drops by 28\% on the evaluation set of {\datasetname}. 
Furthermore, 
the 7\% dropping of ARMSE and 7\% dropping of NME on the subset of BP4D demonstrates the generalization ability of {\datasetname}.
%

%





\section{Conclusion}
In this paper, we introduce a new 3D face dataset, \datasetname, which is composed of over 24,000 synchronous 2D image/3D data pairs with affluent annotations and diversely distributed attributes.
Taking advantage of {\methodname}, we construct a new 3DMM, {\datasetname}, which covers larger shape space and can better model various expressions than previous state-of-the-art 3DMMs.
We conduct massive experiments to train and evaluate state-of-the-art 3D face reconstruction methods on our dataset.
The experiment results reveal the limitations of previous datasets and show that our data can facilitate the models in overcoming these drawbacks to generate more realistic 3D face meshes.
In addition to 3D face reconstruction, our {\datasetname} can also facilitate other research tasks such as 3D face generation and face analysis.

\clearpage
{\small
	\bibliographystyle{ieee}
	\bibliography{3d_face}
}
\end{document}